\documentclass[10pt,twocolumn,letterpaper]{article}

\usepackage{cvpr}
\usepackage{times}
\usepackage{epsfig}
\usepackage{graphicx}
\usepackage{amsmath}
\usepackage{amssymb}

\usepackage{float}
\usepackage{subfigure}
\usepackage{multirow}
\usepackage{booktabs}
\usepackage[breaklinks=true,bookmarks=false]{hyperref}

\cvprfinalcopy 

\def\cvprPaperID{6887} 

\ifcvprfinal\fi
\begin{document}

\title{Customizable Architecture Search for Semantic Segmentation\thanks{{\small This work was performed at JD AI Research.}}}

\author{Yiheng Zhang $^{\dag}$, Zhaofan Qiu $^{\dag}$, Jingen Liu$^{\S}$, Ting Yao $^{\ddag}$, Dong Liu $^{\dag}$, and Tao Mei $^{\ddag}$ \\
         $^{\dag}$ University of Science and Technology of China, Hefei, China\\
         $^{\ddag}$ JD AI Research, Beijing, China~~~~~~~~$^{\S}$ JD AI Research, Mountain View, USA\\
{\tt\small \{yihengzhang.chn, zhaofanqiu, jingenliu, tingyao.ustc\}@gmail.com}\\
{\tt\small dongeliu@ustc.edu.cn, tmei@live.com}
}

\maketitle
\thispagestyle{empty}

    \begin{abstract}
    In this paper, we propose a Customizable Architecture Search (CAS) approach to automatically generate a network architecture for semantic image segmentation. The generated network consists of a sequence of stacked computation cells. A computation cell is represented as a directed acyclic graph, in which each node is a hidden representation (i.e., feature map) and each edge is associated with an operation (e.g., convolution and pooling), which transforms data to a new layer. During the training, the CAS algorithm explores the search space for an optimized computation cell to build a network. The cells of the same type share one architecture but with different weights. In real applications, however, an optimization may need to be conducted under some constraints such as GPU time and model size. To this end, a cost corresponding to the constraint will be assigned to each operation. When an operation is selected during the search, its associated cost will be added to the objective. As a result, our CAS is able to search an optimized architecture with customized constraints. The approach has been thoroughly evaluated on Cityscapes and CamVid datasets, and demonstrates superior performance over several state-of-the-art techniques. More remarkably, our CAS achieves 72.3\% mIoU on the Cityscapes dataset with speed of 108 FPS on an Nvidia TitanXp GPU.
    \end{abstract}

    \section{Introduction}
    Semantic segmentation, which aims at assigning semantic labels to every pixel of an image, is a fundamental topic in computer vision. Leveraging the strong capability of CNNs, which have been widely and successfully applied to image classification \cite{he2016deep,Hu_2018_CVPR,ILSVRC15,simonyan2014very,szegedy2015going}, most state-of-the-art works have made significant progress on semantic segmentation \cite{chen2016deeplab,Chen_2018_ECCV,long2015fully,Peng_2017_CVPR}. To tackle the challenges (e.g., reduced feature resolution and objects at multiple scales) in CNN based semantic segmentation, researchers have proposed various network architectures, such as the application of dilated convolutions \cite{chen2016deeplab,Yu2016ICLR} to capture larger contextual information without losing the spatial resolution, and multi-scale prediction ensemble \cite{xia2016zoom}. Although these methods achieve promising high accuracy, they generally require long inference time due to the complicated networks, which carry huge numbers of operations and parameters.
    \begin{figure}
      \begin{center}
          \includegraphics[width=0.99\linewidth]{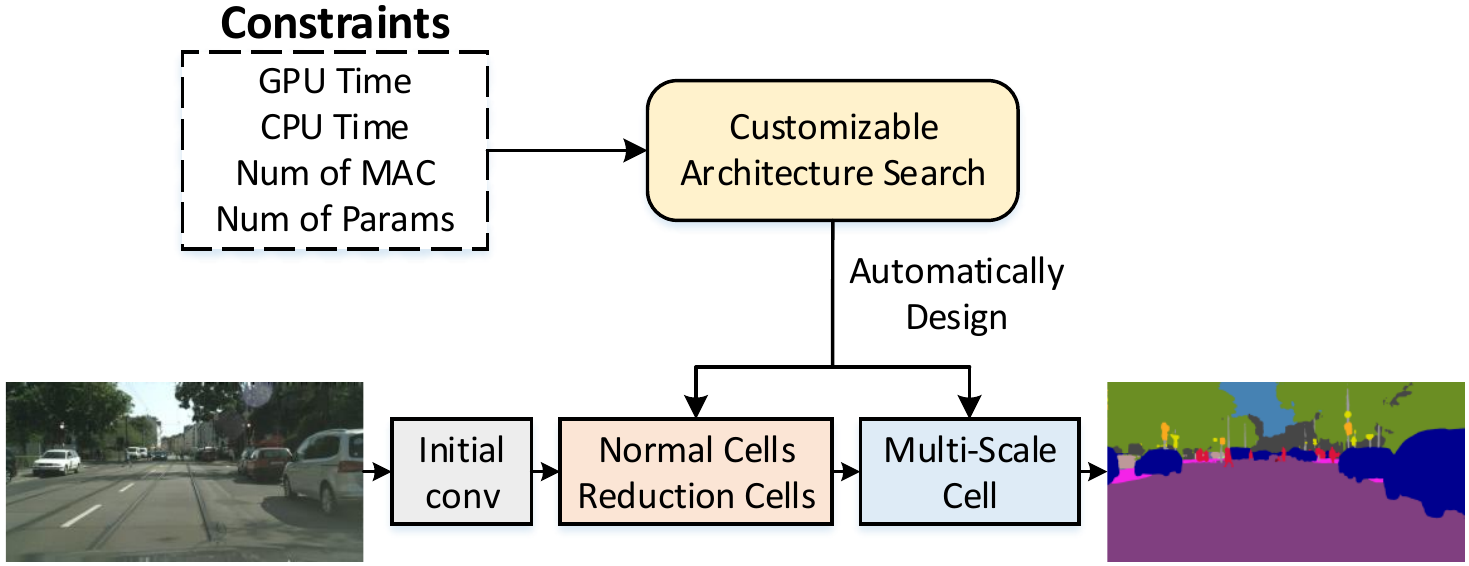}
      \end{center}
      \vspace{-0.1in}
      \caption{\small Our proposed Customizable Architecture Search (CAS) for semantic image segmentation. Given some constraints such as GPU/CPU time and number of parameters, our CAS is able to automatically generate an optimized network which consists of a sequence of stacked computation cells.}
      \label{fig:fig_intro}
      \vspace{-0.15in}
    \end{figure}

    With the increasing need of semantic segmentation on some real-time applications like augmented reality wearables and autonomous driving, there is a high demand for fast semantic segmentation without sacrificing much accuracy, even on a low-power mobile device. Accordingly, some researchers attempt to make a real-time inference by various manually designed strategies including resizing or cropping the input \cite{Zhao_2018_ECCV}, pruning the network channels \cite{SegNet.V}, dropping some stages of the model \cite{paszke2016enet}, multiple scales feature integration \cite{Zhao_2018_ECCV} and spatial-context decoupling \cite{Yu_2018_ECCV}. These designs usually require significant engineering effort of human experts. In addition, they have less flexibility to adjust the inference speed according to the actual dataset and hardware configurations. In other words, it is difficult to find a tradeoff between speed and performance for a specific task. To deal with these issues, we propose a Customizable Architecture Search approach to automatically generate a lightweight network with customized constraints on the availability of computational resource and speed requirements. Our work is inspired by recently proposed solutions to automate the manual process of network design \cite{liu2018darts,Zoph_2018_CVPR}. The successes of these approaches have been demonstrated on some image classification tasks by surpassing the performances of human manually designed architectures \cite{liu2018darts}. Rather than solely pursuing the best performance like \cite{chen2018searching,liu2018darts,Zoph_2018_CVPR}, we aim at searching an appropriate network under the constraints on the computational resource of an application. We call this procedure as Customizable Architecture Search (CAS). To the best of our knowledge, our CAS is the first effort to automatically generate network architectures for semantic segmentation given some constraints in real applications.

    Figure \ref{fig:fig_intro} illustrates an overview of the proposed CAS for semantic segmentation. The proposed lightweight network consists of a couple of initial convolutions followed by sequentially stacked computation cells including both reduction cell and normal cell in the backbone network. A computation cell is a directed acyclic graph, which forms the building block of the learned network. The CAS aims at searching for an optimized architecture to achieve high-quality feature maps. To further recover the loss of spatial information during feature map learning, a multi-scale cell is attached to the backbone network to fuse multiple scales information. CAS jointly learns the architecture of the cells as well as the associated weights. The same type of cells share an identical architecture but with different weights. By relaxing the search space to be continuous, we employ the differential architecture search \cite{liu2018darts} to solve our CAS objective. As a result, the network search can be optimized with respect to a validation set by gradient descent.

    The proposed CAS has been thoroughly evaluated on Cityscapes~\cite{Cordts2016Cityscapes} dataset and promising results have been obtained. To compare with state-of-the-art approaches, we generate architectures constrained by GPU time and evaluate them on Cityscapes~\cite{Cordts2016Cityscapes} and CamVid~\cite{brostow2008segmentation} datasets. The results exceed the state-of-the-art approaches in term of both performance and inference speed.

    \section{Related Work}
    \textbf{CNN based Semantic Image Segmentation.} Inspired by the success of CNN on visual recognition \cite{he2016deep,Hu_2018_CVPR,qiu2017deep,qiu2017learning,ILSVRC15,simonyan2014very,szegedy2015going}, recently researchers have proposed various CNN based approaches for semantic segmentation. The typical way of applying CNNs to segmentation is through patch-by-patch scanning \cite{farabet2013learning,pinheiro2014recurrent}. The fully convolutional network (FCN) \cite{long2015fully} is proposed for semantic segmentation to exploit the high learning capacity of CNNs. It enables spatial dense prediction and efficient end-to-end training. Following FCN, researchers propose several advanced techniques ranging from cross-layer feature ensemble~\cite{ghiasi2016laplacian,Lin:2017:RefineNet,Pohlen_2017_CVPR,xia2016zoom} to context information exploitation~\cite{chen2016deeplab,chen2017rethinking,Chen_2018_ECCV,liu2015parsenet,Peng_2017_CVPR,qiu2018learning,zhang2018fully,zhao2017pspnet}. The FCN formulation could be further improved by employing post-processing techniques, such as the conditional random field \cite{chen2016deeplab}, to consider label spatial consistency.

    A lot of recent efforts have been made to achieve high-quality segmentation without considering the cost such as inference time. For example, PSPNet \cite{zhao2017pspnet} and DeepLabv3 \cite{chen2017rethinking} have achieved over 81\% of mIoU on Cityscape dataset running with less than 2 FPS, which is far away from real-time. Some works attempt to improve the inference speed by restricting the input resolution \cite{SegNet.V}, pruning the channels of the network \cite{Zhao_2018_ECCV}, dropping stages of the model \cite{paszke2016enet} and utilizing the lightweight networks \cite{treml2016speeding}, while the loss of spatial information and network capacity corrupt the dense prediction of semantic segmentation. To remedy the information loss, experienced experts have designed network architectures to balance speed and segmentation quality. ICNet \cite{Zhao_2018_ECCV} is proposed to achieve real-time segmentation with a decent performance by employing a cascade network structure and incorporating multi-resolution branches. BiSeNet \cite{Yu_2018_ECCV} decouples the network into a spatial path and a context path, in order to obtain a faster network with a competitive performance of semantic segmentation. Differing from the aforementioned efforts, in this paper we propose the solution of CAS, which automatically generates a lightweight architecture with the best tradeoff between speed and accuracy under some application constraints.

    \textbf{Network Architecture Search}. The target of architecture search is to automatically design network architectures tailored for a specific task. The sequential model-based optimization~\cite{liu2018progressive} is proposed to guide the searching by learning a surrogate model. The reinforcement learning based methods~\cite{pham2018efficient, Zoph_2018_CVPR}, which train a controller network to generate neural architectures, are proposed to obtain state-of-the-art performances on the tasks of image classification and natural language processing. Instead of treating the architecture search as a black-box optimization problem over a discrete domain, differentiable architecture search (DARTS) \cite{liu2018darts}, which searches architectures in a continuous space, is presented to make the architecture be optimized by gradient descent and achieve competitive performance using fewer computational resources.

    Our work is inspired by \cite{liu2018darts,Zoph_2018_CVPR}. Unlike these methods, however, our work attempts to achieve a good tradeoff between system performance and the availability of the computational resource. In other words, our algorithm is optimized with some constraints from real applications. We notice that the recent DPC work \cite{chen2018searching} is very related to ours. It addresses the dense image prediction problem via searching an efficient multi-scale architecture on the use of performance driven random search \cite{golovin2017google}. Nevertheless, our work is different from \cite{chen2018searching}. First of all, we have different objectives. Instead of targeting high-quality segmentation in~\cite{chen2018searching}, our solution is customizable to search for an optimized architecture which is constrained by the requirements of real applications. The generated architecture tries to keep a balance between the quality and limited computational resource. Secondly, our solution optimizes the architecture of the whole network including both backbone and multi-scale module, while \cite{chen2018searching} focuses on multi-scale optimization. Finally, our method employs a lightweight network, which costs much less training time as compared to that of \cite{chen2018searching}.

    \section{Customizable Architecture Search}
    As shown in Figure \ref{fig:fig_intro}, given the customized constraints in semantic segmentation task, the proposed CAS searches for a computation cell (e.g., normal/reduction cell, and multi-scale cell, which are represented as directed acyclic graphs as depicted in Figure \ref{fig:DAG}) as the building block for an optimized network. Unlike the previous work \cite{liu2018darts}, CAS not only searches for effective operations for a cell, but also considers the cost of choosing these operations. Namely, each operation has an associated cost being selected. As a result, the objective of architecture search is to generate a network that minimizes the following function:
    \begin{equation}
        \small
        \label{object_function}
        \begin{aligned}
            \mathcal{L}_{val} + \lambda\mathcal{L}_{cost}~,
        \end{aligned}
    \end{equation}
    where $\mathcal{L}_{val}$ is the loss on validation dataset, $\mathcal{L}_{cost}$ is the cost associated with the network, and $\lambda$ is the tradeoff controller. To solve this objective, following~\cite{liu2018darts}, we optimize the architecture of the computation cell by using gradient descent. Figure \ref{fig:DAG} illustrates an illustration of generating an architecture with and without constraints. To make this section self-contained, we first discuss the differentiable architecture search of \cite{liu2018darts} in a general form in subsection~\ref{das}. We then describe how to perform the customizable optimization for semantic segmentation in subsection~\ref{cas-opt} , and detail the search space for network backbone and multi-scale cell in subsection~\ref{backbone-search} and~\ref{mscell-search}, respectively.

    \subsection{Differentiable Architecture Search\label{das}}
\begin{figure}
      \centering
      \includegraphics[width=0.47\textwidth]{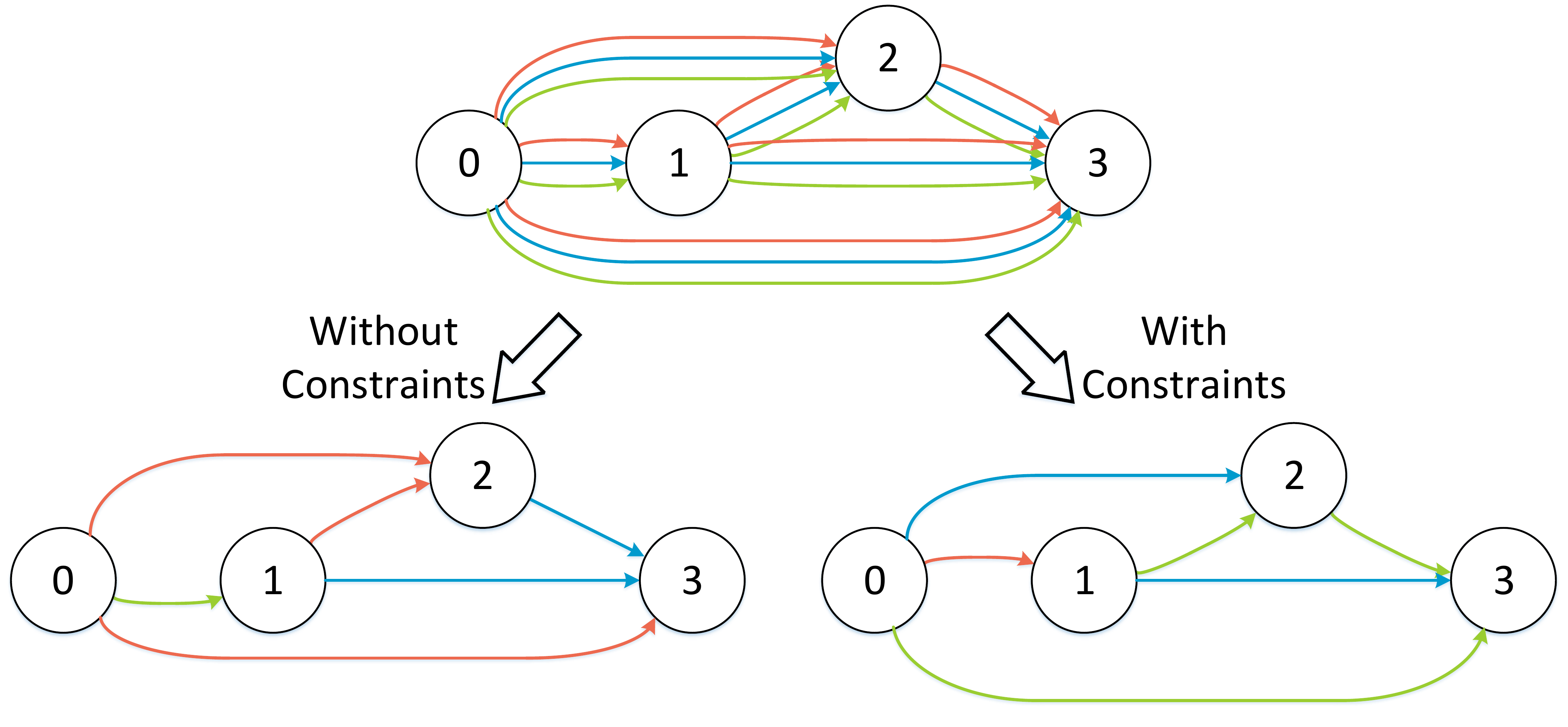}
      \caption{\small An illustration of generating a computing cell with/without constraints. Each edge represents one operation between two nodes. The top graph shows many candidate operations existing between nodes, and each candidate operation has its own cost. The red edge denotes a heavy cost, and the green one has a light cost. Without considering constraints, the search may generate a costly architecture (bottom left) for better performance, while our CAS outputs an architecture with light cost (bottom right).}
      \label{fig:DAG}
      \vspace{-0.10in}
    \end{figure}

    A computation cell is a directed acyclic graph (DAG) as shown in Figure \ref{fig:DAG}. The graph has an ordered sequence of $N$ nodes, represented as $\mathcal{N}=\{x^{(i)}|i=1,\dots,N\}$, where $x^{(i)}$ denotes the feature map in a convolutional network. The transformation from $x^{(i)}$ to $x^{(j)}$ is represented as an operation $o^{(i,j)}(\cdot)$, which corresponds to a directed edge in the graph. Each computation cell has two input nodes (i.e., outputs of the previous two layers) and one output node (i.e., the concatenation of the intermediate nodes in the cell). Specifically, an intermediate node is calculated as:
    \begin{equation}
        \small
        \label{intermediate_node}
        \begin{aligned}
        {x^{(j)}=\sum_{i<j}o^{(i,j)}(x^{(i)})}~,
        \end{aligned}
    \end{equation}
    where $x^{(i)}$ is a node coming before $x^{(j)}$ in the cell. Hence, the problem of architecture search is equivalent to learning the operation on each edge in DAG.

    To make the search space continuous, a weighted combination of all candidate operations is utilized as the transformation on the directed edge as follows:
    \begin{equation}
        \small
        \label{intermediate_node}
        \begin{aligned}
        \bar o^{(i,j)}(x)=\sum_{o\in \mathcal{O}}Softmax(\alpha_{o}^{(i,j)})o(x)~,
        \end{aligned}
    \end{equation}
    where $o(\cdot)$ is an operation in the operation candidate set $\mathcal{O}$ of size $N_o$, and $\alpha_{o}^{(i,j)}$ is a learnable score of the operation $o(\cdot)$. The vector $\alpha^{(i,j)}\in\mathbb{R}^{N_o}$ represents the scores of all candidate operations on the edge from $x^{(i)}$ to $x^{(j)}$. Then the cell architecture is denoted as $\alpha=\{\alpha^{(i,j)}\}$, which is a set of vectors for all edges. Now the architecture search could be formulated as finding $\alpha$ to minimize the validation loss $\mathcal{L}_{val}(w'(\alpha),\alpha)$, where $w'(\alpha)$ is the parameters of the operations. The parameters are obtained by minimizing the training loss, formulated as $w'(\alpha)=argmin_{w}\mathcal{L}_{train}(w,\alpha)$. Accordingly, a cell could be optimized by adjusting $\alpha$ via gradient descent.

    Since the variation of $\alpha$ leads to the recomputation of $w'(\alpha)$ by minimizing $\mathcal{L}_{train}(w,\alpha)$, the optimization procedure could be approximately performed by alternately optimizing weight parameters $w$ and cell architecture $\alpha$ with gradient descent steps. In particular, for the parameter update step $k$, $w_{k-1}$ is moved to $w_k$ according to the gradient $\triangledown_w\mathcal{L}_{train}(w_{k-1}, \alpha_{k-1})$, and the architecture is updated to minimize the validation loss:
    \begin{equation}
        \small
        \label{validation_loss}
        \begin{aligned}
        \mathcal{L}_{val}(w_k-\xi \triangledown_w\mathcal{L}_{train}(w_{k}, \alpha_{k-1}),\alpha_{k-1})~,
        \end{aligned}
    \end{equation}
    where $\triangledown_w\mathcal{L}_{train}(w_k, \alpha_{k-1})$ is a virtual gradient step of $w_k$ and $\xi$ is the step's learning rate.
    After optimizing the architecture of the computation cell encoded as $\alpha$ via gradient descent, each operation combination $\bar o^{(i,j)}$, which locates on the directed edge from $x^{(i)}$ to $x^{(j)}$ of the DAG, is replaced with the most likely operation candidate according to $\alpha^{(i,j)}$. Then $k$ strongest predecessors of each intermediate node are retained, where the strength of an edge is defined as $max(Softmax(\alpha^{(i,j)}))$. The $k$ is set as 2 in the following sections.
    \begin{figure*}
        \begin{center}
            \includegraphics[width=0.99\linewidth]{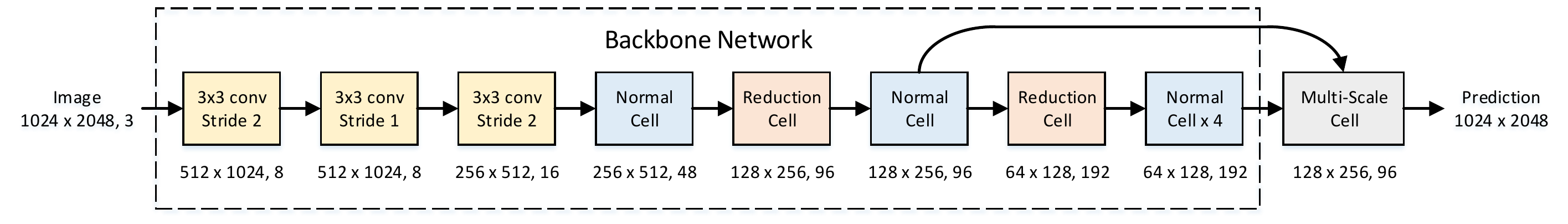}
        \end{center}
        \vspace{-0.10in}
        \caption{\small An overview of our network structure for semantic segmentation. We take $1024\times2048$ input as an example. It consists of two main components: the backbone network on the left followed by the multi-scale cell on the right. The backbone designed for efficient feature extraction begins with three convolutional layers followed by 6 normal cells and 2 reduction cells. The multi-scale cell learns to refine the feature map by integrating accurate spatial information from the second normal cell into the final feature map. Each cell employs the previous two cells' outputs as its inputs.}
        \label{fig:fig2}
        \vspace{-0.10in}
    \end{figure*}
    \subsection{Customizable Optimization \label{cas-opt}}
    As aforementioned, the differentiable architecture search enables an efficient search of high-performance architecture. Nevertheless, considering some practical constraints in real applications, a high-performance architecture is not the only pursuit given limited computational resource. In this section, we propose a constrained architecture search method, which takes a further step forward to discover an appropriate design of the network satisfying customizable constraints. To address the constraints in the architecture search procedure, we associate a cost with each operation, such that whenever an operation is selected, there is a cost for the selection. Hence, the cost of a cell is formulated~as:
    \begin{equation}
        \small
        \label{cell_cost}
        \begin{aligned}
        \mathcal{L}_{cost}=\sum_{j}\sum_{i<j}\sum_{o\in\mathcal{O}}c_{o}Softmax(\alpha_{o}^{(i,j)})~,
        \end{aligned}
    \end{equation}
    where $c_o$ is the cost associated with operation $o(\cdot)$ (Please refer to the implementation details in section~\ref{CAS_implementation} for how to convert constraints to costs). Hence, the architecture is optimized by updating $\alpha$ according to the following gradient:
    \begin{equation}
        \small
        \label{cell_cost_grad}
        \begin{aligned}
        \triangledown_{\alpha}\mathcal{L}_{val}+\lambda\triangledown_{\alpha}\mathcal{L}_{cost}~,
        \end{aligned}
    \end{equation}
    where $\lambda$ is the tradeoff parameter and maintains the balance between the performance and network cost.

    When applying CAS to semantic image segmentation, we employ a network structure as shown in Figure \ref{fig:fig2}. It mainly contains two components: backbone and multi-scale cell, which are built and optimized by CAS separately. Given the input images, the backbone is first utilized to learn feature representations with rich semantics, while the accuracy of pixel-level localization will accordingly drop due to consecutive down-sampling operations. On the other hand, the multi-scale cell learns a refinement structure to recover spatial information from the feature on different stages of the backbone and leads to better predictions for semantic segmentation. The following two sections describe the details of the search for both components, respectively.

    \subsection{Backbone Architecture Search \label{backbone-search}}
    As shown in Figure \ref{fig:fig2}, the backbone network starts with three convolutional layers, followed by eight cells, each of which consists of $N=6$ nodes including the input and output node. The first two nodes of the $i$-th cell are the outputs of the ($i-1$)-th and ($i-2$)-th cells or layers with $1\times1$ convolutions if dimension projection needed. In general, a backbone for image classification contains 5 spatial reduction which results in a feature map of 1/32 size of the original image \cite{he2016deep, simonyan2014very, szegedy2015going}. Different from image classification which focuses on semantic aggregation, the loss of spatial information caused by spatial reduction is more important for semantic segmentation. As such, following \cite{chen2017rethinking}, the spatial resolution of the final feature map is set only 16 times smaller than the input image resolution to balance the spatial density, semantics and expensive computation. In our case, in addition to the first and third convolutional layers of the backbone network with strides of 2, the two reduction cells also serve for down-sampling the feature map. Except for the reduction cells, the other cells are normal cells without reduction. Hence, the searchable architectures of the backbone are represented as $\alpha_{normal}$ and $\alpha_{reduce}$ shared by all normal cells and reduction cells, respectively, but with different weights.

    We draw inspiration from the recent advances in the CNN literatures and collect the operation set $\mathcal{O}_{b}$:
    \begin{table}[H]
        \footnotesize
        \setlength{\tabcolsep}{0.5em} 
        \centering
        \vspace{-0.15in}
        \begin{tabular}{ll}
        $\bullet\;$ identity \\
        $\bullet\;$ 3x3 max pooling                     & $\bullet\;$ 3x3 separable conv, repeat 2 \\
        $\bullet\;$ 3x3 ave pooling                     & $\bullet\;$ 3x3 separable conv, repeat 4 \\
        $\bullet\;$ 3x3 conv                            & $\bullet\;$ 3x3 conv, repeat 2 \\
        $\bullet\;$ 3x3 dilated conv                    & $\bullet\;$ 3x3 dilated conv, repeat 2 \\
        \multicolumn{2}{l}{$\bullet\;$ 2x2 ave pooling stride 2 + 3x3 conv + upsampling}     \\
        \multicolumn{2}{l}{$\bullet\;$ 2x2 ave pooling stride 2 + 3x3 conv repeat 2 + upsampling}
        \end{tabular}
        \vspace{-0.15in}
    \end{table}
    The $\mathcal{O}_{b}$ consists of four types of operations, i.e., non-learned operations, standard convolutions, separable convolutions and pooled convolutions. The identity shortcut \cite{he2016deep}, max pooling and average pooling are non-learned operations. The standard $3\times3$ convolutional layers with optional dilation are widely utilized in the convolutional networks designed for semantic segmentation. The separable convolution proposed in \cite{Chollet_2017_CVPR} is an operation that efficiently balances cost and performance by factorizing the standard convolution into a depthwise convolution and a pointwise convolution. It is worth noting that the separable convolution is often applied at least twice in an operation \cite{liu2018darts,Zoph_2018_CVPR}. In addition to existing operations, we propose the spatial bottleneck operation, namely pooled convolution. This operation applies average pooling with stride 2 on the feature map, followed by $3\times3$ convolutions and finally recovers the resolution of the feature map via bilinear upsampling. Our experiments demonstrate that such operation could effectively enlarge the receptive field and reduce computational cost. Note that we also repeat each weighted operation twice to enlarge the potential capacity of backbone~network.

    \subsection{Multi-Scale Cell Search \label{mscell-search}}
    With an optimized backbone network, the high-quality feature maps learned from images could be obtained and fed into the classifier to generate dense predictions for the images. To further refine feature maps by recovering the spatial information, multi-scale fusion, which aggregates different level features, has been proved to be effective for semantic segmentation \cite{Chen_2018_ECCV,ghiasi2016laplacian,Lin:2017:RefineNet,long2015fully,Peng_2017_CVPR,Yu_2018_ECCV,Zhao_2018_ECCV}. In this paper, we aim at searching a multi-scale cell rather than directly utilizing manually designed architectures. The cell $\alpha_{ms}$ consisting of $N=9$ nodes is heavier than $\alpha_{normal}$ and $\alpha_{reduce}$ in terms of cost. Nevertheless, the cell $\alpha_{ms}$ is only applied once at the end of the network and thus the cost is negligible compared to other cells. In $\alpha_{ms}$, the spatial resolutions of the inputs are firstly aligned by upsampling the smaller one via bilinear interpolation and then independent $1\times1$ convolutions are applied on each directed edge from spatially aligned inputs to intermediate nodes for channel projection. Inspired by the recent works on semantic segmentation, an operation set $\mathcal{O}_{ms}$ is collected specifically as:
    \begin{table}[H]
        \footnotesize
        \setlength{\tabcolsep}{1.0em} 
        \centering
        \vspace{-0.1in}
        \begin{tabular}{ll}
        $\bullet\;$ 3x3 conv, dilation=1           & $\bullet\;$ 3x3 conv, dilation=2 \\
        $\bullet\;$ 3x3 conv, dilation=4           & $\bullet\;$ 3x3 conv, dilation=8 \\
        $\bullet\;$ 15x1 then 1x15 conv            & $\bullet\;$ 25x1 then 1x25 conv \\
        $\bullet\;$ 8x8 residual SPP               & $\bullet\;$ 16x16 residual SPP \\
        $\bullet\;$ 24x24 residual SPP             & $\bullet\;$ identity \\
        \end{tabular}
        \vspace{-0.15in}
    \end{table}
    Three types of operations, i.e., standard convolutions, spatial decomposed convolutions and residual spatial pyramid pooling, are included in $\mathcal{O}_{ms}$. Convolutional layers with multiple dilations could effectively capture multi-scale information \cite{Chen_2018_ECCV}. The spatial decomposed convolution with large kernel size enables densely connections within a large region in the feature map and embeds rich context information in each location with less computational cost than general convolution with large kernel \cite{Peng_2017_CVPR}. To provide contextual scenery prior to the feature map, the residual spatial pyramid pooling (SPP) with different window sizes is explored. Inside of each window, an average pooling is performed followed by an $1\times1$ convolution to encode the contextual information. The spatial resolution of the encoded context, which is combined with input feature map as residual value, is recovered by bilinear upsampling.

    \section{Implementation}
    \subsection{Customizable Architecture Search \label{CAS_implementation}}
    We utilize the gradient in Eq.(\ref{cell_cost_grad}) to update the $\alpha$ in CAS, The $\triangledown_{\alpha}\mathcal{L}_{val}$ could be derived from Eq.(\ref{validation_loss}) as:
    \begin{equation}
        \small
        \label{validation_grad}
        \begin{aligned}
        \triangledown_{\alpha}\mathcal{L}_{val}(w',\alpha)-\xi\triangledown_{\alpha,w}^{2}\mathcal{L}_{train}(w,\alpha)\triangledown_{w'}\mathcal{L}_{val}(w', \alpha)~,
        \end{aligned}
    \end{equation}
    where $w'=w-\xi\triangledown_w\mathcal{L}_{train}(w,\alpha)$. The weight parameters $w$ are updated by the virtual gradient step. For ease of optimization, an approximation of Eq.(\ref{validation_grad}) is applied and the gradient of architecture could be represented as $\triangledown_{\alpha}\mathcal{L}_{val}(w,\alpha)$ with respect to the case of $\xi=0$ on the assumption that $\alpha$ and $w$ are independent.

    Given the candidate operation set $\mathcal{O}$, to evaluate the cost $c_o$, we firstly measure the cost $c_o'$ of the whole network whose cells only consist of $o(\cdot)$, and $c_o$ is computed as $c_o=c_o'-c_{id}'$, where $c_{id}$ denotes the cost of the network whose operations of cells are replaced by ``identity''. The cost could be defined according to the constraints, e.g., GPU / CPU inference time, number of parameters and number of multiply-accumulate operations (MAC). In order to characterize the lack of concatenation between two nodes in the computation cell, a special ``None'' operation is appended to $\mathcal{O}$ during the optimization but this operation is excluded in the decision of the final architecture.
    \begin{figure*}
        \centering
        \subfigure[\small $1k$ iters, mIoU=62.4\%, time=14.1ms]{
          \label{fig:fig4:a}
          \includegraphics[width=0.32\textwidth]{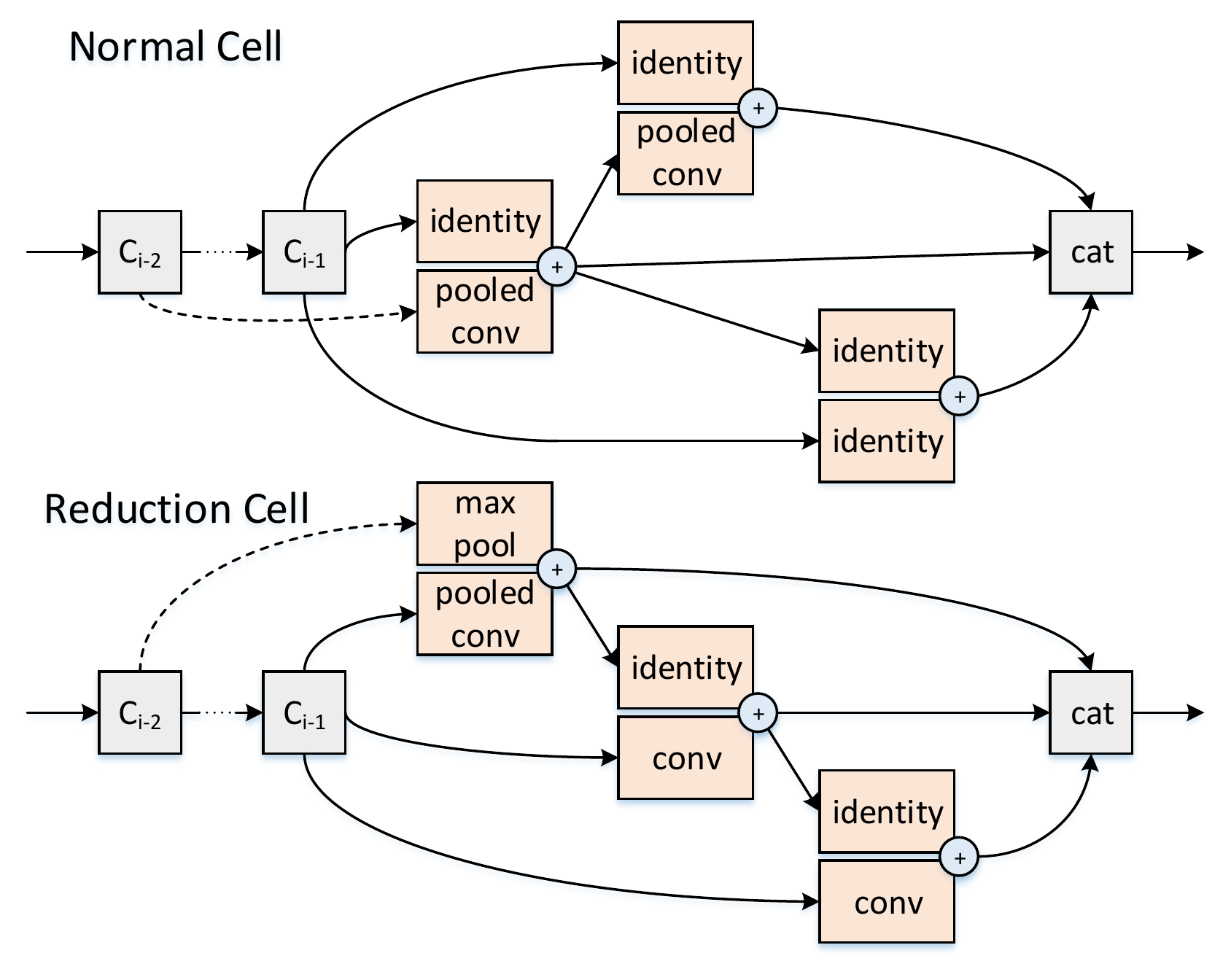}}
        \subfigure[\small $5k$ iters, mIoU=64.6\%, time=22.4ms]{
          \label{fig:fig4:b}
          \includegraphics[width=0.32\textwidth]{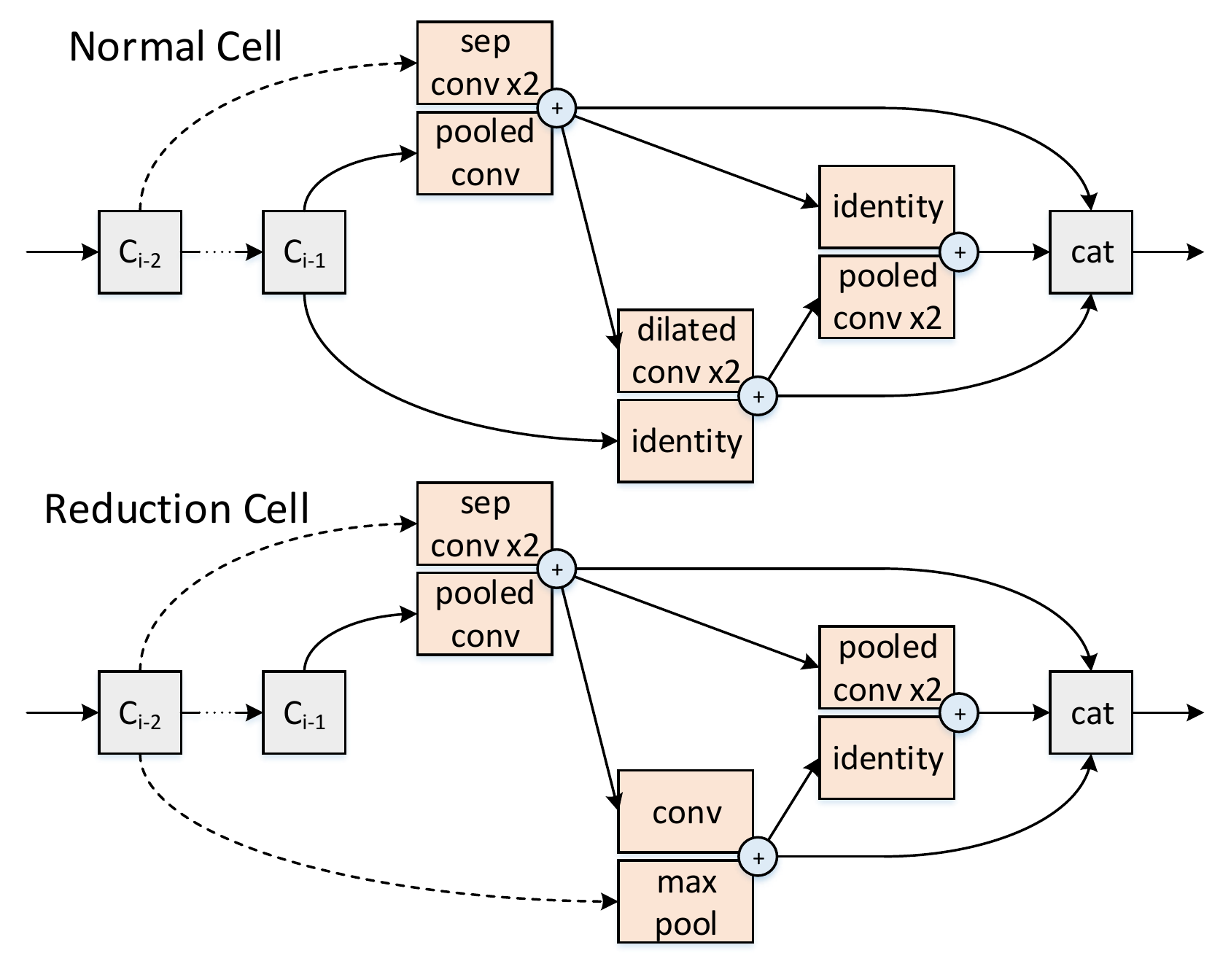}}
        \subfigure[\small $15k$ iters, mIoU=68.1\%, time=23.8ms]{
          \label{fig:fig4:c}
          \includegraphics[width=0.32\textwidth]{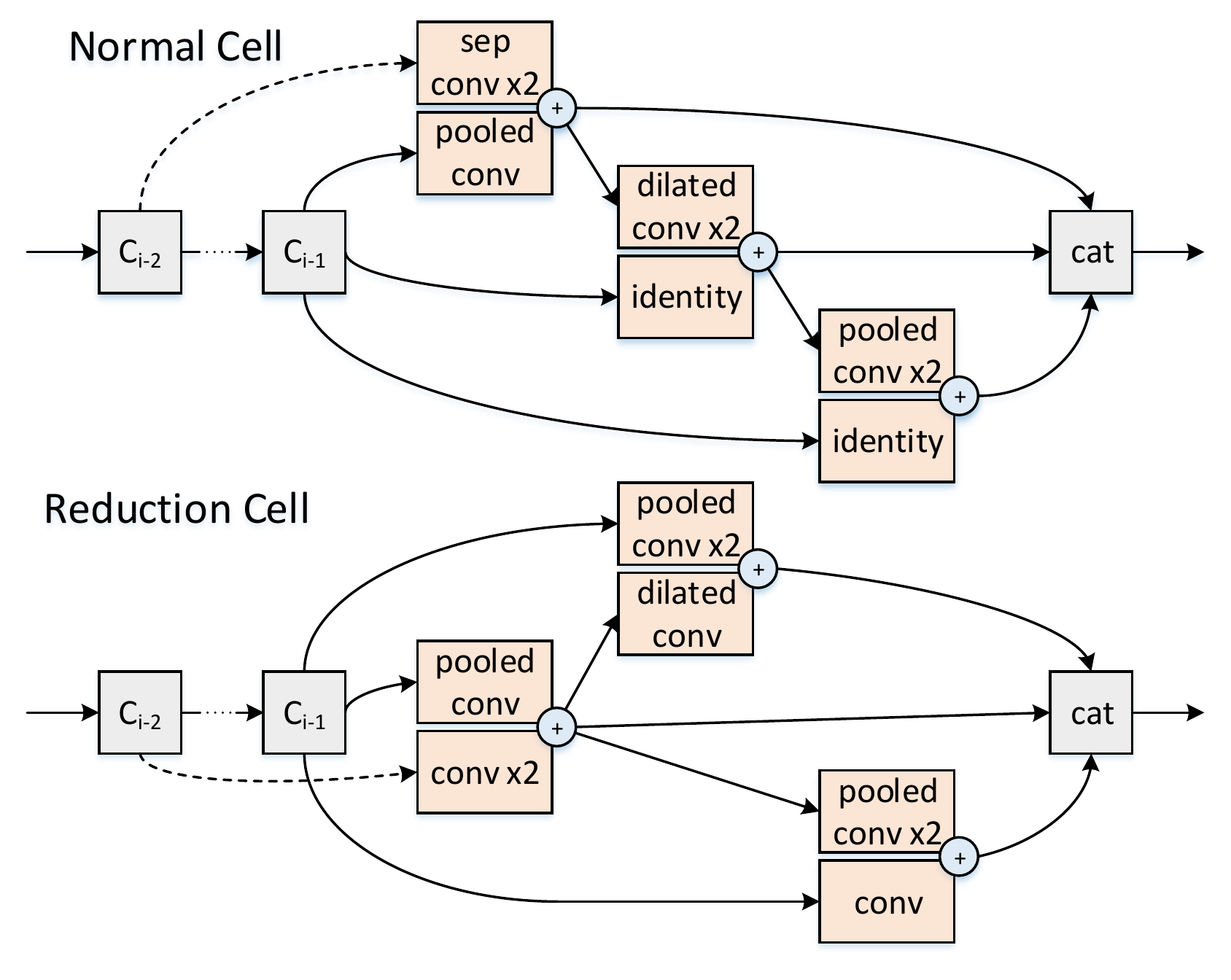}}
        \caption{\small Examples of the normal cell and reduction cell during CAS procedure with the GPU Time constraint. The performance of the network is consistently increased from 62.4\% to 68.1\% with the increase of iterations, and the inference time converges to 23.8ms.}
        \label{fig:fig4}
        \vspace{-0.15in}
     \end{figure*}

    \subsection{Semantic Segmentation}
    In our implementations, we search the architectures of the backbone network and multi-scale cell separately. The backbone network architecture is firstly determined according to $\alpha_{normal}$ and $\alpha_{reduce}$ which are both optimized by CAS on the task of semantic segmentation. Then we utilize ImageNet ILSVRC12 dataset \cite{ILSVRC15} to pre-train the backbone network from the scratch. With the ImageNet pre-trained weights, the multi-scale cell is appended at the top of the backbone. The architecture is fixed by $\alpha_{ms}$ after the procedure of CAS. The whole network initialized with the ImageNet pre-trained weights in backbone, is finally optimized on semantic segmentation.

    \subsection{Training Strategy}
    Our proposal is implemented on Caffe \cite{jia2014caffe} framework with CUDNN, and mini-batch stochastic gradient descent algorithm is exploited to optimize the model. In the search procedure of CAS, the initial learning rate is 0.005. We exploit the ``poly'' learning rate policy with power fixed to 0.9. Momentum and weight decay are set to 0.9 and 0.0005, respectively. The batch size is 16. The maximum iteration number is $15k$. To evaluate the architecture generated by CAS, we train the whole network for $90k$ iterations. The rest hyper-parameters are the same as those in the search procedure of CAS.

    \section{Experiments}
    In all experiments, the Intersection over Union (IoU) per category and mean IoU over all the categories are used as the performance metric. The resolution of the input image is $1024\times2048$, and the GPU/CPU inference time is reported on one Nvidia GTX 1070 GPU card and Intel i7~8700 CPU, respectively, unless otherwise stated.

    \subsection{Datasets}
    We conduct a thorough evaluation of CAS on Cityscapes \cite{Cordts2016Cityscapes}, one popular benchmark for semantic understanding of urban street scenes. It contains high-quality pixel-level annotations of 5,000 images collected in street scenes from 50 different cities. The image resolution is $1024\times2048$. Following the standard protocol in segmentation task \cite{Cordts2016Cityscapes}, 19 semantic labels are used for evaluation. The training, validation, and test sets contain 2975, 500, and 1525 images, respectively. An additional set of 23,473 coarsely annotated images are also available in this dataset. In our evaluation, the training set is further split into two groups, which play the roles of ``training set'' (1599 images from 9 cities) and ``validation set'' (1376 images from another 9 cities) in architecture search, respectively. Note that the original validation set or test set is never used for architecture search.

    Moreover, we also evaluate the merit of CAS on the CamVid dataset, which is a standard scene parsing dataset. There are five video sequences in total with resolution up to $720\times960$. The sequences are densely labeled at one frame per second with 11 class labels. We follow the training/testing split in \cite{brostow2008segmentation}, with 468/233 labeled frames in the dataset for training/testing.
    \begin{figure*}
        \centering
        \includegraphics[width=0.90\textwidth]{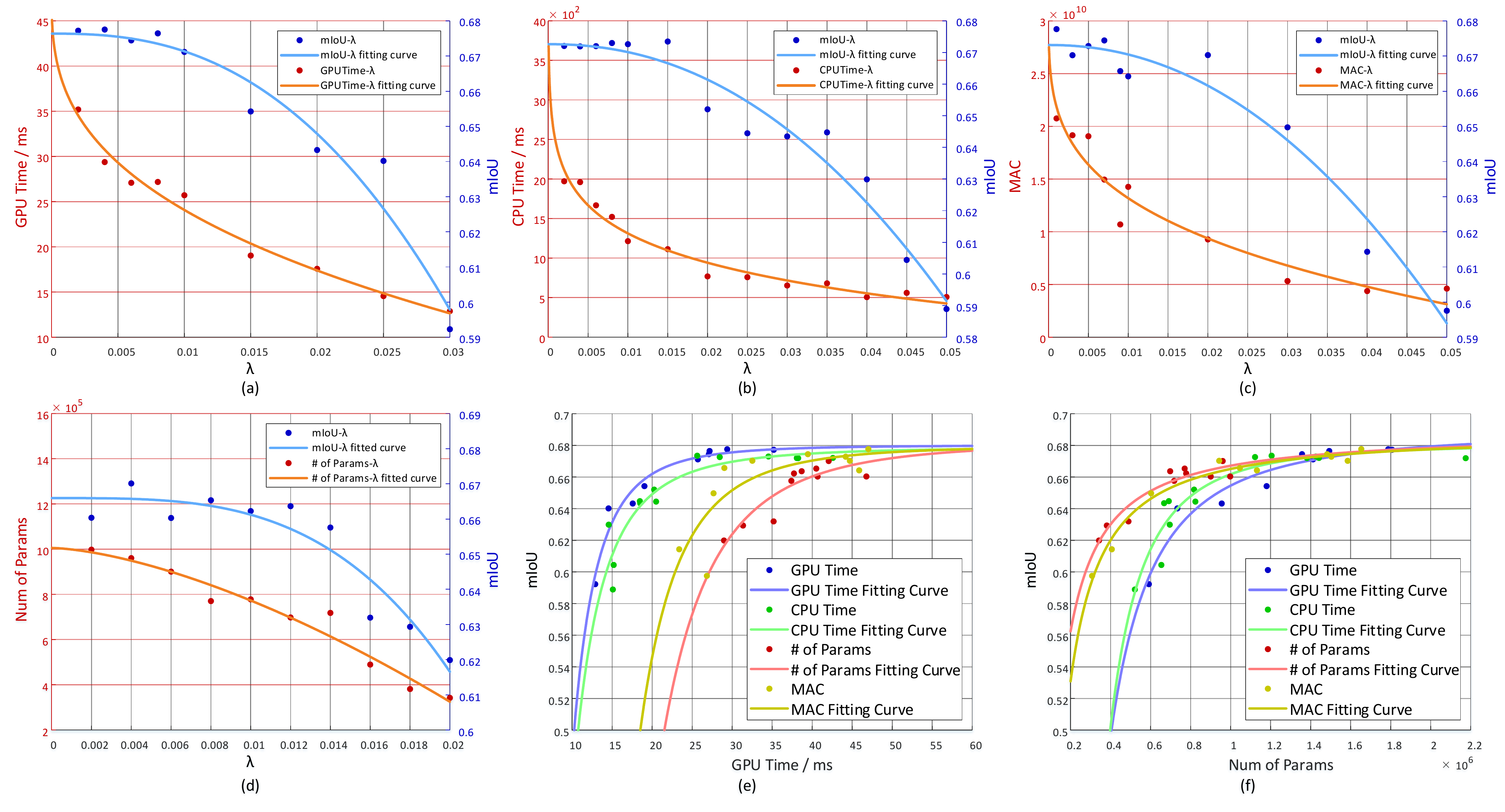}
        \caption{\small (a) GPU Time-$\lambda$ and mIoU-$\lambda$ curves under the constraint of GPU time. (b) CPU Time-$\lambda$ and mIoU-$\lambda$ curves under the constraint of CPU time. (c) MAC-$\lambda$ and mIoU-$\lambda$ curves under the constraint of MAC. (d) Num of Params-$\lambda$ and mIoU-$\lambda$ curves under the constraint of number of parameters. (e) mIoU-GPUTime curves under four constraints. (f) mIoU-Num of Params curves under four constraints. Better viewed in original color pdf.}
        \label{fig:fig5}
        \vspace{-0.20in}
     \end{figure*}
    \subsection{Evaluation of CAS}
    \textbf{Architecture search by CAS.} First, we conduct experiments to explore the evolution procedure of the architecture optimization given some constraints. The architecture search is performed on Cityscapes training set from the scratch and the searched architectures are evaluated on Cityscapes validation set. Figure \ref{fig:fig4} illustrates the architecture evolution of a normal cell and a reduction cell during the CAS optimization given the constraint on GPU time. Let us look at how a normal cell architecture changes during CAS optimization, which attempts to reach a tradeoff between network performance and GPU time. As shown in Figure \ref{fig:fig4:a}, at the beginning of optimization, the cell selects the most lightweight operation ``identity'' and ``pooled conv'', which is able to immediately decrease the network computation by reducing the spatial resolution. As a result, the inference of the network is fast with relatively low mIoU, i.e., 62.4\%@14.1ms. When iterating the search process $5k$ times, heavy operations (e.g., separable convolution and dilated convolution) are selected in pursuit of better performance by sacrificing some inference time (from 14.1ms (a) to 22.4ms (b)) as shown in Figure \ref{fig:fig4:b}. The search converges after $15k$ iterations to reach a cell in Figure \ref{fig:fig4:c} and no extra heavy operations are employed after $5k$ iterations in our observations. The results indicate that CAS could optimize cells well with the constraints during architecture search. The whole search procedure of cells verifies our design that the performance and customized constraints of the network could be automatically balanced by CAS.

    \textbf{CAS with different constraints.} Then, we conduct another group of experiments to demonstrate the effectiveness of CAS. More specifically, we examine the impact of the tradeoff parameter $\lambda$ towards a balance between segmentation performance and constraint costs. All the experiments are evaluated on Cityscapes validation set with networks trained on the training set from the scratch. The experiment on each setting is repeated five times, and the average values are reported. Figure \ref{fig:fig5}(a)$\sim$\ref{fig:fig5}(d) depicts results under constraints of GPU time, CPU time, MAC, and Number of Parameters, respectively. The blue and red points/curves in the figure illustrate the mIoU and cost of networks given different $\lambda$ values, and the curves are fit to the points utilizing 2-terms power function. All the experiments consistently show that the network cost decreases rapidly with the increase of $\lambda$, resulting in the drop of the performance. Please also note that a small increment of $\lambda$ could lead to a significantly reduced cost but without notably sacrificing the performance, especially when $\lambda$ is relatively small. In other words, we could expect an affordable network whose performance is not much worse than that of the costly ones.

    \begin{figure}
      \begin{center}
          \includegraphics[width=0.97\linewidth]{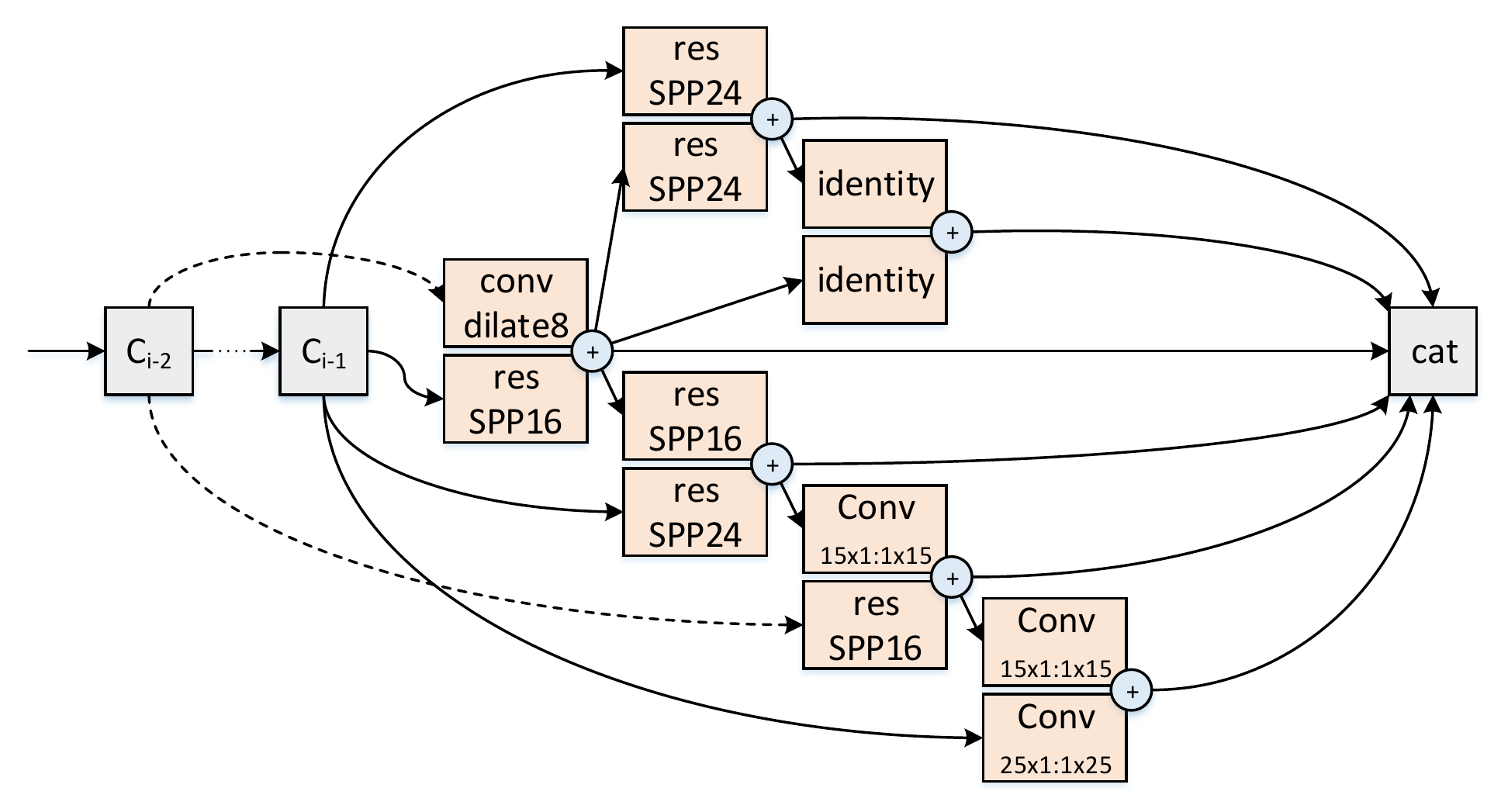}
      \end{center}
      \caption{\small The architecture of the multi-scale cell.}
      \label{fig:fig_mscell}
      \vspace{-0.25in}
    \end{figure}

    Next, we turn to compare the network design of CAS with respect to different constraints. Figure \ref{fig:fig5}(e) and \ref{fig:fig5}(f) shows the mIoU performances when utilizing GPU time and number of parameters as the measure of cost under each constraint, respectively. In the two figures, each curve depicts the performances of networks which are generated by CAS with the corresponding constraint. For instance, the blue and green curve in Figure \ref{fig:fig5}(e) represents the performances of the networks optimized with constraints of GPU time and CPU time, respectively. As expected, optimizing networks when setting the alignment of constraint and the computation on cost will lead to better performance. Specifically, capitalizing on the constraint of GPU time constantly exhibits an mIoU boost over other constraints when computing cost on GPU time. Similarly, when the cost is calculated on number of parameters, the networks with the constraint of number of parameters achieve the best mIoU. The results indicate the flexibility of our CAS for architecture search with customizable constraints.

    \begin{table}[]
        \centering
        \small
        \caption{\small Evaluation of pre-training and multi-scale cell.}
        \begin{tabular}{ll|c|c} \hline
        \multicolumn{2}{l|}{\textbf{Method}}                                            &  \textbf{mIoU (\%)} &  \textbf{Time (ms)}       \\ \hline
        \multicolumn{2}{l|}{CAS-GT}                                                     &  68.1               &  23.8               \\ \hline
        \multicolumn{2}{l|}{+ImageNet Pre-train}                                        &  70.4               &  23.8               \\ \hline
        \multicolumn{1}{l|}{\multirow{4}{*}{+MSC}}  & PSP\cite{zhao2017pspnet}          &  71.5               &  26.5               \\
        \multicolumn{1}{l|}{}                       & ASPP\cite{chen2017rethinking}     &  72.9               &  33.2               \\
        \multicolumn{1}{l|}{}                       & ASPP+\cite{Chen_2018_ECCV}        &  73.9               &  56.9               \\
        \multicolumn{1}{l|}{}                       & MSCell                            &  \textbf{74.0}      &  \textbf{29.2}      \\ \hline %
        \end{tabular}
        \label{tab:mscell}
        \vspace{-0.20in}
    \end{table}

    \textbf{Evaluation of the multi-scale cell.} The multi-scale cell is employed to recover the spatial information loss caused by the downsampling operations in the backbone network. Here, we study how the multi-scale cell influences the overall performance. Let CAS-GT be the best backbone network searched by CAS under the constraint of GPU time and $\lambda$ = 0.01. The multi-scale cell is placed at the top of CAS-GT. The architecture of the multi-scale cell searched by CAS, which is denoted as MSCell, is illustrated in Figure \ref{fig:fig_mscell}. As the most frequently selected operation, the residual pyramid pooling benefits from its capability of gathering the context information from large regions and preserving fine spatial information. Table \ref{tab:mscell} details the mIoU and GPU time of CAS-GT with and without the multi-scale cell. In our case, ImageNet pre-training successfully boosts up the mIoU performance from 68.1\% to 70.4\% without additional inference time. Utilizing multi-scale cells (MSC) at the top of ImageNet pre-trained CAS-GT could further increase the mIoU of the network. Particularly, PSP\cite{zhao2017pspnet}, ASPP\cite{chen2017rethinking} and ASPP+\cite{Chen_2018_ECCV}, which are manually designed multi-scale cells, obtain 1.1\%, 2.5\% and 3.5\% performance gains with extra 2.7ms, 9.4ms and 33.1ms inference time, respectively. Compared to the manually designed ones, our MSCell leads to an mIoU increase of 3.6\% and the mIoU performance reaches 74.0\% with only 5.4ms additional inference~time.

    \subsection{Real-time Semantic Segmentation}
    In this section, we validate CAS with the configuration of CAS-GT plus MSCell on the scenario of real-time semantic segmentation. The architecture search is optimized with the constraint of GPU time. We run all the inferences on an Nvidia TitanXp GPU card and calculate the frame per second (FPS) for all the methods. For fair comparisons, we measure the speed of the methods based on our implementations if the original speed was reported on different GPUs.

    \begin{table}[t]
        \centering
        \small
        \caption{\small mIoU and inference FPS on Ciytscapes validation (\textit{val}) and test (\textit{test}) sets. The mIoU and inference FPS of our method are given on the downsampled images with resolution $768\times1536$.}
        \begin{tabular}{l|c|c|c} \hline
        \multirow{2}{*}{\textbf{Method}}                 & \multicolumn{2}{c|}{\textbf{mIoU (\%)}}                                 & \multirow{2}{*}{\textbf{FPS}}\\ \cline{2-3}
                                                         & \multicolumn{1}{c|}{\textit{val}}  & \multicolumn{1}{c|}{\textit{test}} &                              \\ \hline
        FCN-8s \cite{long2015fully}                      &  -                                &  65.3                               &  4.4                         \\
        Dilation10 \cite{Yu2016ICLR}                     &  68.7                             &  67.1                               &  0.7                         \\
        PSPNet \cite{zhao2017pspnet}                     &  -                                &  81.2                               &  1.3                         \\
        DeepLabv3 \cite{chen2017rethinking}              &  -                                &  81.3                               &  1.3                         \\ \hline
        SegNet \cite{SegNet.V}                           &  -                                &  57.0                               &  33.0                        \\
        ENet \cite{paszke2016enet}                       &  -                                &  58.3                               &  78.4                        \\
        SQ \cite{treml2016speeding}                      &  -                                &  59.8                               &  21.7                        \\
        ICNet \cite{Zhao_2018_ECCV}                      &  67.7                             &  69.5                               &  37.7                        \\
        ICNet \cite{Zhao_2018_ECCV} (+coarse)            &  -                                &  70.6                               &  37.7                        \\
        BiSeNet-Xception39 \cite{Yu_2018_ECCV}           &  69.0                             &  68.4                               &  105.8                       \\
        BiSeNet-Res18 \cite{Yu_2018_ECCV}                &  74.8                             &  74.7                               &  65.5                        \\ \hline
        CAS-GT+MSCell                                             &  71.6                             &  70.5                               &  108.0                       \\
        CAS-GT+MSCell (+coarse)                                   &  72.5                             &  72.3                               &  108.0                       \\ \hline
        \end{tabular}
        \label{tab:cityscapes}
        \vspace{-0.10in}
    \end{table}

    \textbf{Results on Cityscapes.} We evaluate CAS-GT+MSCell on Cityscapes validation and test sets. The validation set is included for training when submitting our network to online Cityscapes server and evaluating the performance on official test set. Following \cite{Yu_2018_ECCV}, we scale the resolution of the image from $1024\times2048$ to $768\times1536$, and measure the speed and mIoU without other evaluation tricks. Both the performance and FPS comparisons are summarized in Table \ref{tab:cityscapes}. Overall, our CAS-GT+MSCell is the fastest among all the methods. Compared to BiSeNet-Xception39 \cite{Yu_2018_ECCV} which is as fast as ours, CAS-GT+MSCell leads to an mIoU performance boost of 2.1\% on the test set. Compared to the methods designed for high-speed semantic segmentation such as ENet \cite{paszke2016enet}, SQ \cite{treml2016speeding} and ICNet \cite{Zhao_2018_ECCV}, CAS-GT+MSCell achieves faster inference and makes performance improvement over them by 12.2\%, 10.7\% and 1.0\%, respectively. The results demonstrate the effectiveness of our CAS for balancing performance and constraints. When additionally leveraging coarse annotations of Cityscapes, CAS-GT+MSCell yields the mIoU of 72.3\% on test set.

    \textbf{Results on CamVid.} To validate the transferability of learnt architectures, we perform the experiments on CamVid with the cells searched on Cityscapes for real-time semantic segmentation. Note that we merely transfer the architectures of CAS-GT+MSCell but train the weights on CamVid. Table \ref{tab:camvid} details the comparisons of both performance and inference time on CamVid test set. The input resolution is $720\times960$. In particular, our CAS-GT+MSCell surpasses the best competitor BiSeNet-Res18 by 2.5\% in mIoU. More importantly, the inference speed of CAS-GT+MSCell achieves 169 FPS, which is very impressive. The results basically verify the merit of CAS from the aspect of network generalization.

    \begin{table}[t]
        \centering
        \small
        \caption{\small mIoU and inference FPS on CamVid test set. The mIoU and inference FPS of our method are given on the original images with resolution $720\times960$.}
        \begin{tabular}{l|c|c} \hline
        \textbf{Method}                                  & \textbf{mIoU (\%)}                   & \textbf{FPS}   \\ \hline
        Dilation8 \cite{Yu2016ICLR}                      &  65.3                                &  6.5           \\ 
        PSPNet50  \cite{zhao2017pspnet}                  &  69.1                                &  6.8           \\ \hline
        SegNet    \cite{SegNet.V}                        &  55.6                                &  29.4          \\ 
        ENet      \cite{paszke2016enet}                  &  51.3                                &  61.2          \\ 
        ICNet     \cite{Zhao_2018_ECCV}                  &  67.1                                &  34.5          \\ 
        BiSeNet-Xception39\cite{Yu_2018_ECCV}            &  65.6                                &  -             \\ 
        BiSeNet-Res18\cite{Yu_2018_ECCV}                 &  68.7                                &  -             \\ \hline
        CAS-GT+MSCell                                             &  \textbf{71.2}                       &  \textbf{169.0}\\ \hline
        \end{tabular}
        \label{tab:camvid}
        \vspace{-0.20in}
    \end{table}

    \section{Conclusion}
    In this paper, we propose an approach to automatically generate a network architecture for semantic image segmentation. Unlike some previous approaches, which require huge efforts from human experts to manually design a network, our approach utilizes a lightweight framework, and automatically searches for optimized computation cells which are the building blocks of the network. In addition, our CAS takes the constraints of real applications into account when optimizing the architecture. As a result, ours is able to seek a good balance between segmentation performance and available computational resource. Experiments on both Cityscapes and CamVid datasets demonstrate the advantages over other state-of-the-art approaches.

    \textbf{Acknowledgments.} This work was supported in part by the Strategic Priority Research Program of the Chinese Academy of Sciences under Grant XDB06040900.

{\small
\bibliographystyle{ieee_fullname}
\bibliography{egbib}
}

\end{document}